\documentclass[
]{ceurart}

\usepackage{listings}

\usepackage{multirow}
\usepackage[utf8]{inputenc}
\usepackage[english, russian]{babel}
\newcommand{\includecapchar}[1]{\includegraphics[width=0.5em]{#1}}
\newcommand{\includelowchar}[1]{\includegraphics[width=0.4em]{#1}}

\newcommand{\CQ}{\includecapchar{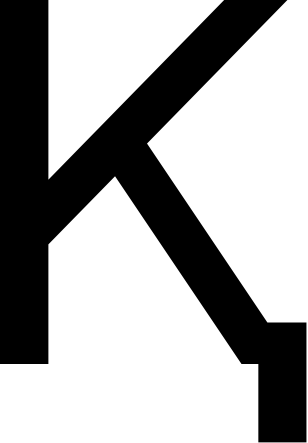}}
\newcommand{\Cq}{\includelowchar{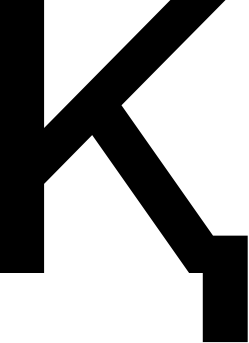}}
\newcommand{\CH}{\includecapchar{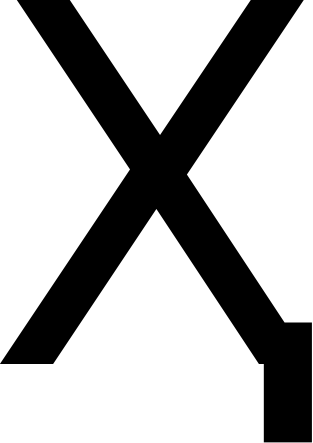}}
\newcommand{\Ch}{\includelowchar{unicode-chars/h-capital.pdf}}
\newcommand{\CG}{\includecapchar{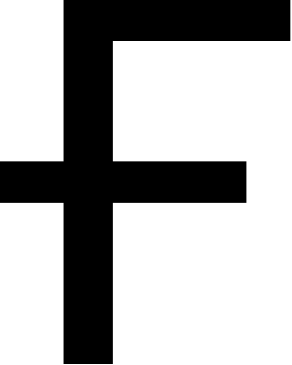}}
\newcommand{\Cg}{\includelowchar{unicode-chars/g-capital.pdf}}

\DeclareUnicodeCharacter{2205}{$\emptyset$}
\DeclareUnicodeCharacter{02BB}{`} 
\DeclareUnicodeCharacter{02BC}{'} 

\begin{document}
\selectlanguage{english}
\copyrightyear{2021}
\copyrightclause{Copyright for this paper by its authors.
  Use permitted under Creative Commons License Attribution 4.0
  International (CC BY 4.0).}

\conference{The International Conference on Agglutinative Language Technologies as a challenge of Natural Language Processing (ALTNLP), June 6, 2022, Koper, Slovenia}

\title{A machine transliteration tool between Uzbek alphabets}

\author[1]{Ulugbek Salaev}[%
orcid=0000-0003-3020-7099,
email=ulugbek0302@gmail.com
]
\author[2]{Elmurod Kuriyozov}[%
orcid=0000-0003-1702-1222,
email=e.kuriyozov@udc.es
]
\author[2]{Carlos G\'omez-Rodr\'iguez}[%
orcid=0000-0003-0752-8812,
email=carlos.gomez@udc.es,
url=http://www.grupolys.org/~cgomezr
]

\address[1]{ Urgench State University, Department of Information Technologies, 14, Kh.Alimdjan str, Urgench city, 220100, Uzbekistan
}
\address[2]{ Universidade da Coru\~na, CITIC, Grupo LYS, Depto. de Computaci\'on y Tecnologías de la Información, Facultade de Inform\'atica, Campus de Elvi\~na, A Coru\~na 15071, Spain
}

\begin{abstract}
Machine transliteration, as defined in this paper, is a process of automatically transforming written script of words from a source alphabet into words of another target alphabet within the same language, while preserving their meaning, as well as pronunciation. The main goal of this paper is to present a machine transliteration tool between three common scripts used in low-resource Uzbek language: the old Cyrillic, currently official Latin, and newly announced New Latin alphabets. The tool has been created using a combination of rule-based and fine-tuning approaches. The created tool is available as an open-source Python package, as well as a web-based application including a public API. To our knowledge, this is the first machine transliteration tool that supports the newly announced Latin alphabet of the Uzbek language.

\end{abstract}

\begin{keywords}
  transliteration \sep
  uzbek language \sep
  natural language processing \sep
  low-resource language
\end{keywords}

\maketitle

\section{Introduction}
The term transliteration is ambiguous, as it refers to two similar tasks of Natural Language Processing (NLP), which differ according to their either inter-language or intra-language nature. More specifically, a transliteration can be described as a process of representing words from one language using the alphabet of another language \cite{arbabi1994algorithms}, while the other use of the term stands for the act of transforming words from one alphabet into another alphabet within the same language \cite{birnbaum1967transliteration}. We take the latter case as our goal in this work, and present a method for transforming words between three equally-important alphabets of the low-resource Uzbek language.

\textbf{Uzbek language}  (native: \textit{O‘zbek tili}) is a low-resource, highly-agglutinative language with null-subject and null-gender characteristics from the Karluk branch of the Turkic language family. It is an official language of Uzbekistan, with more than 30 million speakers inside and around the country, making it the second most widely spoken language among Turkic languages (right after Turkish language)\footnote{More about Uzbek language: \url{https://en.wikipedia.org/wiki/Uzbek_language}}.

The Cyrillic alphabet had been in use for a long time for Uzbek language, until it was replaced with Latin script in 1993\footnote{Law of the Republic of Uzbekistan ``On the introduction of the Uzbek alphabet based on the Latin script'' (September 2, 1993 year, reg. number: 931-XII): \url{https://lex.uz/docs/-112286}.} 
(with a reformation in 1995\footnote{On Amendments to the Law of the Republic of Uzbekistan ``On Introduction of the Uzbek Alphabet Based on the Latin Script'' (May 6, 1995 year, reg. number: 71-I): \url{https://lex.uz/docs/-116158}.}), 
which is still an official alphabet. 
The use of both Cyrillic and Latin alphabets is equally popular in all areas of written language (law, books, web, media, etc.) even these days. Availability of texts in two writing systems make it harder and costlier for NLP researchers and practitioners to work on the language, such as by limiting the amount of collected data for a specific alphabet, or by creating a need to develop language resources and models for both alphabets. Furthermore, there is a new reformation\footnote{Resolution of the Cabinet of Ministers of the Republic of Uzbekistan ``On measures to ensure a gradual transition to the Uzbek alphabet based on the Latin script.'' (February 10, 2021 year, reg. number: 61): \url{https://lex.uz/uz/docs/-5281850}.} 
that has been introduced to change all the existing digraphs and replace them with diacritical signs\footnote{More about alphabets used in Uzbek language: \url{https://www.omniglot.com/writing/uzbek.htm}.}, so every letter in the alphabet would be written with only a single character. Throughout this paper, we refer to this reformed Latin alphabet as ``New Latin'' alphabet.

Considering the existence of three distinctive alphabets currently in use in Uzbek language, we propose a methodology to perform the task of transliteration between those three alphabets, which is a combination of basic rule-based character-mapping, more sophisticated cross-alphabet specific rules, as well as fine-tuning approaches. 
Although there are some available web tools that offer transliteration between Cyrillic and Latin alphabets for Uzbek, none of them offer neither an open source code, nor an Application Programming Interface (API) for integration with other tools. Moreover, the only one tool with a good quality from Savodxon project\footnote{\url{https://savodxon.uz/}} is commercial, and the free ones are not practical enough to be used, due to a bad implementation.
In this paper, we also present a publicly available Python code\footnote{\url{https://github.com/UlugbekSalaev/UzTransliterator}} for research integration, together with a web-based tool\footnote{\url{https://nlp.urdu.uz/?menu=translit}} that also includes an API, which is, to our knowledge, the first ever transliteration tool between all three alphabets. 


\section{Related work}
One of the very early mentions of machine transliteration was raised by Nida and Taber \cite{nida1969theory}, stating that a problem of ``untranslatability'' arises when an exact equivalence of meanings is required in translation, rather than a comparative equivalence, so they referred to transliteration to tackle the issue. In early mentions, transliteration was described as a process of representing words from one language using the alphabet of another language, as part of machine translation \cite{arbabi1994algorithms}. Later on, it also has been used for similar purposes, but with intra-language perspective, describing it as a conversion of words from one written script to another one within the same language \cite{birnbaum1967transliteration,alam2017sequence}.

Instances of early works on transliteration can be Arabic-English names transliteration using a combination of a rule-based system with neural networks \cite{arbabi1994algorithms}, and Japanese-English using finite state transducers \cite{knight1997machine}. Both approaches dealt with phonetic representations of words, which were replaced by a spelling-based approach to achieve higher results, as in the case of the Arabic-English model of \cite{al2002machine}.
Later modern approaches to transliteration include models with long short-term memories (LSTM) \cite{alam2017sequence}, and recurrent neural networks (RNN) \cite{le2019low}, which perform equally well. Combination of old rule-based approaches with recent deep-learning methods improves the quality, according to a comparative study \cite{najafi2018comparison}. 

Transliteration between Cyrillic and Latin alphabets of Uzbek language has been done by Mansurovs \cite{mansurov2021uzbek}, who used a data-driven approach, by aligning words and training a decision-tree classifier.
Among some other NLP work that has been done on low-resource Uzbek language so far, there are a morphological analyzer \cite{matlatipov2009representation}, WordNet type synsets \cite{agostini2021uzwordnet}, Uzbek stopwords dataset \cite{madatov2022automatic}, sentiment analysis and text classification \cite{kuriyozov2019building,rabbimov2020investigating,rabbimov2020multi}, cross-lingual word-embeddings \cite{kuriyozov2020cross}, as well as a pretrained Uzbek language model based on the BERT architecture \cite{mansurov2021uzbert}.

\section{Methodology}
To check the accuracy of our tool, we collected text from the spelling dictionary of Uzbek language \cite{togayev1999ozbek}. This dictionary is a printed resource that contains about 14K commonly-used words of Uzbek language in Latin and Cyrillic variants. We did not include multiword expressions, because we used word-level evaluation to check the performance analysis, and using them by splitting into single words would create duplications. After also removing words that could not be successfully digitalized using OCR, we ended up with around 9600 words to use in our experiments.

Although the dictionary size is limited, it includes words that are prone to spelling errors between Cyrillic and Latin. Since there is no publicly available data for the New Latin alphabet yet, we transliterated those words from Latin to New Latin, then manually checked resulting words, correcting them where necessary. Manual correction was only possible within our resources thanks to the fact that the majority of words stayed the same as in Latin, we focused only on words that changed their form.

The methodology used in this work is very similar to the work from Mansurovs \cite{mansurov2021uzbek}, but we extend it by adding the New Latin alphabet. Additionally, instead of training a classifier, we rely on string replacement techniques for the sake of simplicity and speed.
Following are the steps followed by the tool, and the steps that need more detail are explained separately afterwards:
\begin{enumerate}
    \item \verb|Tokenization:| Feeding text from the source alphabet as string buffer, and splitting into tokens;
    \item \verb|Replacement of exceptional words:| Checking each token to see if it is or contains a word from the exceptional words dataset (excluding punctuations, emojis, or unrecognized characters), if so, replacing it with its target version;
    \item  \verb|Replacement using rules:|Going through a set of mapping rules specific to the pair of alphabets and conversion direction that were designed to use where one-to-one character mapping does not apply. Technically, each rule consists of a simple regular exrpression that looks for a specific sub-string (usually one to three character long), and replaces it with desired sub-string(either empty, one or more characters long);
    \item  \verb|Character-mapping:|Replacing the rest of characters from source alphabet to the target one using one-to-one mapping. This can also be made by very simple regular expression that replaces one character with another in a string;
    \item  \verb|Re-uniting:|Merging resulting tokens that contain target alphabet characters back again, and returning them as a whole string. 
\end{enumerate}

\selectlanguage{russian}
\subsection{Replacement of exceptional words}
This is the step we came up with after applying a fine-tuning approach to the created tool. There are words that cannot be transliterated using a rule-based approach. Only one-directional transliteration (like from Cyrillic to Latin) may be possible, but it could fail in the opposite direction (like from Latin to Cyrillic). To solve this issue, we extracted words from the collected data that did not provide the same output when transliterated and back-transliterated between different combinations. So far, there are 233 words with their form in all three alphabets that are stored in the tool as an exceptional words database. Some examples of such words can be seen in Table \ref{table:exception-words}. One interesting insight about those words is that they are mostly loan words from Russian language, and there is usually a change when converting Cyrillic letters \textit{ц}, \textit{ь} (phonetic glottal stop), and \textit{я}. Although this process was done after the tool's creation, it is required that this step has to be applied before any further conversion steps are applied.
\selectlanguage{english}
\begin{table}[!ht]
\centering
\caption{Some examples from the exceptional words database where rule-based transliteration does not apply.}
\label{table:exception-words}
\selectlanguage{russian}
\begin{tabular}{llll}
\multicolumn{1}{c}{\textbf{Latin}} & \multicolumn{1}{c}{\textbf{Cyrillic}} & \multicolumn{1}{c}{\textbf{New Latin}} & \multicolumn{1}{c}{\textbf{English}} \\ \hline
aksent       & акцент      & aksent       & accent               \\
budilnik     & будильник   & budilnik     & alarm clock          \\
batalyon     & батальон    & batalyon     & batalion            \\
feldsher     & фельдшер    & feldşer      & paramedic            \\
fransuz      & француз     & fransuz      & french               \\
intervyu     & интервью    & intervyu     & interview            \\
koeffitsient & коэффициент & koeffitsient & coefficient          \\
korrupsiya   & коррупция   & korrupsiya   & corruption           \\
kuryer       & курьер      & kuryer       & courier              \\
medalyon     & медальон    & medalyon     & medallion            \\
oktabr       & октябрь     & oktabr       & october              \\
pavilyon     & павильон    & pavilyon     & pavilion             \\
porshen      & поршень     & porşen       & piston               \\
shpatel      & шпатель     & şpatel       & scraper (putty knife) \\
cherepitsa   & черепица    & çerepitsa    & roof tile (shingle)  \\ \hline
\end{tabular}
\end{table}

\selectlanguage{russian}
\subsection{Character-mapping}
Steps 3 and 4 of the conversion deal with mapping characters from source alphabet to the target one. Although the majority of letters are replaced in a straightforward manner, the remaining characters require set of pairwise rules based on the alphabets involved, and the direction of the conversion. A general idea of  conversion between alphabets is given in Table \ref{table:character-mapping}.

Throughout the process, we found out that some conversion rules are not as straightforward as expected. There is a problem with handling a single character uppercase letter when converting  to a digraph letter in other alphabet. For instance, if we convert Cyrillic uppercase letters \textit{Ш} and \textit{Ю} into \textit{SH} and \textit{YU} (respectively) in Latin, an error like these happen: "\textit{Шўрва}"->"\textit{S\textbf{H}o'rva}" (soup), or "\textit{Юлдуз}"->"\textit{Y\textbf{U}lduz}" (star); But if we convert it into  \textit{Sh} and \textit{Yu}, then an error with acronyms occurs like these: "\textit{АҚШ}"->"\textit{AQS\textbf{h}}" (USA), or "\textit{ЮНЕСКО}"->"\textit{Y\textbf{u}NESKO}" (UNESCO). A solution to this kind of problem is to consider surrounding letters when performing conversion.

Another complicated situation with mapping rules is the phonetic glottal stop  (native: \textit{Tutuq belgisi}), which is also part of an alphabet in Uzbek language. There are some words that a glottal stop appears in its Cyrillic form and is omitted in its Latin form. For instance:  "\textit{факул\textbf{ь}тет}"->"\textit{fakultet}" (faculty), or "\textit{кал\textbf{ь}ций}"->"\textit{kalsiy}" (calcium). The problem with this omission is twofold: The algorithm has to be taught whether to omit it or not, also when these words are transliterated back to Cyrillic, the glottal stop has to appear out of nowhere. A solution to this kind of problem is to include this kind of words in the exceptional words list.
\selectlanguage{english}
\begin{table}[!ht]
\centering
\caption{Character-level mapping between alphabets for transliteration. \textbf{Cyr.} stands for Cyrillic alphabet, \textbf{Lat.} stands for Latin alphabet, and \textbf{NewLat.} stands for New Latin alphabet. ∅ denotes an empty string. Highlighted rows indicate a complex mapping, where one character from source alphabet is mapped to either two or zero characters from target alphabet. The character at the very end of the table is called a \textit{phonetic glottal stop} (native: \textit{Tutuq belgisi}), and although it is not a real letter, still it is considered a part of the Uzbek alphabet.}
\label{table:character-mapping}
\selectlanguage{russian}
\begin{tabular}{ccccccccccc}
\textbf{Cyr.} &
  \textbf{Lat.} &
  \textbf{NewLat.} &
  \multirow{13}{*}{} &
  \textbf{Cyr.} &
  \textbf{Lat.} &
  \textbf{NewLat.} &
  \multirow{13}{*}{} &
  \textbf{Cyr.} &
  \textbf{Lat.} &
  \textbf{NewLat.} \\ \cline{1-3} \cline{5-7} \cline{9-11} 
А а & A a       & A a       &  & Л л & L l & L l &  & \CH~\Ch   & H h       & H h       \\ \cline{1-3} \cline{5-7} \cline{9-11} 
Б б & B b       & B b       &  & М м & M m & M m &  & \textbf{Ц ц}   & \textbf{Ts/S ts/s} & \textbf{Ts/S ts/s} \\ \cline{1-3} \cline{5-7} \cline{9-11} 
В в & V v       & V v       &  & Н н & N n & N n &  & Э э   & E e       & E e       \\ \cline{1-3} \cline{5-7} \cline{9-11} 
Г г & G g       & G g       &  & О о & O o & O o &  & \textbf{Ю ю}   & \textbf{Yu/u yu/u} & \textbf{Yu/u yu/u} \\ \cline{1-3} \cline{5-7} \cline{9-11} 
Д д & D d       & D d       &  & П п & P p & P p &  & \textbf{Я я}   & \textbf{Ya/A ya/a} & \textbf{Ya/A ya/a} \\ \cline{1-3} \cline{5-7} \cline{9-11} 
\textbf{Е е} & \textbf{E/Ye e/ye} & \textbf{E/Ye e/ye} &  & \CQ~\Cq & Q q & Q q &  & \textbf{Ў ў}   & \textbf{Oʻ oʻ}     & \textbf{Ō ō}       \\ \cline{1-3} \cline{5-7} \cline{9-11} 
\textbf{Ё ё} & \textbf{Yo yo}     & \textbf{Yo yo}     &  & Р р & R r & R r &  & \textbf{\CG~\Cg}   & \textbf{Gʻ gʻ}     & \textbf{Ḡ ḡ}       \\ \cline{1-3} \cline{5-7} \cline{9-11} 
Ж ж & J j       & J j       &  & С с & S s & S s &  & \textbf{Ш ш}   & \textbf{Sh sh}     & \textbf{Ş ş}       \\ \cline{1-3} \cline{5-7} \cline{9-11} 
З з & Z z       & Z z       &  & Т т & T t & T t &  & \textbf{Ч ч}   & \textbf{Ch ch}     & \textbf{Ç ç}       \\ \cline{1-3} \cline{5-7} \cline{9-11} 
И и & I i       & I i       &  & У у & U u & U u &  & \textbf{Нг нг} & \textbf{Ng ng}     & \textbf{Ñ ň}       \\ \cline{1-3} \cline{5-7} \cline{9-11} 
Й й & Y y       & Y y       &  & Ф ф & F f & F f &  & \textbf{ъ}     & \textbf{ʼ/∅}       & \textbf{ʼ/∅}       \\ \cline{1-3} \cline{5-7} \cline{9-11} 
К к & K k       & K k       &  & Х х & X x & X x &  &       &           &           \\ \cline{1-3} \cline{5-7}
\end{tabular}
\end{table}

\begin{table}[!ht]
    \centering
    \caption{Micro-averaged F1 scores of word level transliteration process between alphabets. The direction of the transliteration is from the alphabet shown in the row to the alphabet shown in the column.}
    \label{table:f1-scores}
    \begin{tabular}{l|c|c|c}
        \multicolumn{1}{c|}{\textbf{Alphabets}} & \textbf{Latin} & \textbf{Cyrillic} & \textbf{New Latin} \\ \hline
        \textbf{Latin}    & -    & 0.89 & 0.94 \\ \hline
        \textbf{Cyrillic} & 0.90 & -    & 0.92 \\ \hline
        \textbf{New Latin} & 0.93 & 0.92 & -   
    \end{tabular}
\end{table}

\section{Results}
The created tool has been analysed using the collected parallel text data for all three alphabets, and comparing the tool's output for each word with the actual expected output. We have calculated micro-averaged F1 scores of each conversion using the \textit{metrics} module of scikit-learn\footnote{\url{https://scikit-learn.org/0.15/modules/classes.html##module-sklearn.metrics}}. F1-scores are calculated at the word level (i.e., by considering words that the system transliterates correctly or incorrectly). Table \ref{table:f1-scores} shows the results between each pair and each direction.

Although the analysis has been done using very limited amount of data, it gives us some insights about the tool's performance: The best performing pair is Latin->New Latin conversion (0.94 F1 score) due to the reason that there are only five letters that change during conversion with no exceptional cases (to our best knowledge), and those errors that still occur are only because of the problem with handling the abbreviations.The worst performing pair is Latin->Cyrillic (0.89 F1 score), likely due to many conversion rules to consider, plus many exceptional cases.
Furthermore, It is also possible to see that transliteration to and from the New Latin alphabet performs better than any other alphabets do, which can be explained by the minimum number of conversion rules required compared to its counterparts. More specifically, Transliteration between New Latin and Latin would require only 5 specific conversion rules (and no exceptional cases), and 6 rules (plus exceptional cases) between New Latin and Cyrillic, while the same process would require 11 rules (and exceptional cases) from a transliteration between Latin and Cyrillic alphabets.

\begin{figure}[!ht]
    \centering
    \frame{\includegraphics[width=0.9\columnwidth]{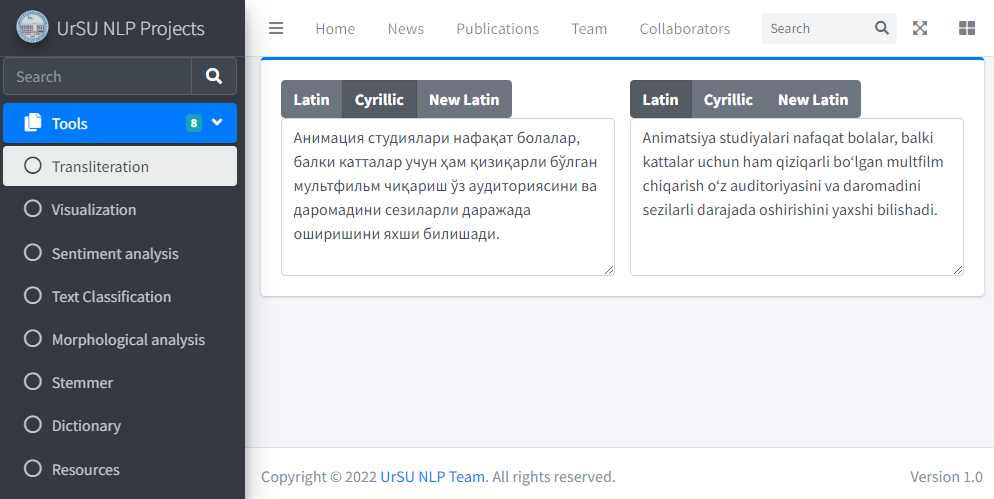}}
    \caption{Web-interface of the created transliteration tool.}
    \label{fig:web-interface}
\end{figure}

The Python tool created for this work is openly-accessible, and also can be easily installed, using the following command that is popular for the Python community:
\begin{lstlisting}
pip install UzTransliterator
\end{lstlisting}
The user interface of the created web tool can be seen in Figure \ref{fig:web-interface}.
There is also a public API based on this tool, and more detailed information about it can be found at the project's GitHub repository. 

\subsection{Discussion}
Although the created tool is practical enough to be used for transliteration, there are some certain cases we still have to consider and improve on the go:
\begin{itemize}
\item Our database of exceptional cases (a result of fine-tuning approach) contains only lemmas of words, and due to the highly agglutinative nature of the Uzbek language, words mostly appear as inflections and derivations. For this reason we have to either store their root forms or add syntactic knowledge to handle all possible forms of lemmas; 
\item New loan words and proper nouns adopted from other languages might not produce expected output, thus we have to keep updating the database of exceptional cases;
\item We dealt with legal text properly written in Uzbek language, which is not always the case with user-generated text. Especially, there is a big deal of inconsistency in writing \textit{o'} and \textit{g'} letters in the currently official alphabet due to the use of apostrophe, which comes in many ways, such as \textit{oʻ,o`,o'}, and \textit{gʻ,g`,g'} forms respectively;
\item Due to the lack of texts created in the New Latin alphabet, we worked only with manually created text, which is very limited and requires more analysis as the coverage starts to enlarge. 
\end{itemize}


\section{Conclusion}
In this paper, we presented a Python code, a web tool, and an API created for the low-resource Uzbek language that performs machine transliteration between two popularly used Cyrillic and Latin alphabets, as well as a newly reformed version of the Latin alphabet which, according to the governmental decree, all legal texts will have been completely adapted to by year 2023.
We have also shown the cases of alphabet-specific problems related to the transliteration between those three scripts that do not allow for a simple character mapping, including ongoing attempts to tackle user-input related issues.

Our future work will be to strengthen the output quality of the current tool by implementing more mapping rules, user input cleaning techniques, as well as integrating a pretrained neural language model that can handle unseen cases. Furthermore, we hope to be able to make a pipeline that can perform useful NLP tasks for Uzbek language, such as tokenization, POS tagging, morphological analysis, and parsing in a foreseen future.

\begin{acknowledgments}
This work has received funding from ERDF/MICINN-AEI (SCANNER-UDC, PID2020-113230RB-C21), from Xunta de Galicia (ED431C 2020/11), and from Centro de Investigación de Galicia ``CITIC'', funded by Xunta de Galicia and the European Union (ERDF - Galicia 2014-2020 Program), by grant ED431G 2019/01. Elmurod Kuriyozov was funded for his PhD by El-Yurt-Umidi Foundation under the Cabinet of Ministers of the Republic of Uzbekistan.
\end{acknowledgments}

\bibliography{main}


\end{document}